# The Self-Driving Portfolio:
# Agentic Architecture for Institutional Asset Management

**September 21, 2026**

**Andrew Ang** is co-founder of Tau Balance and Altbridge, and an adjunct professor at Columbia University in New York, NY. www.andrewangphd.com (Corresponding author)

**Nazym Azimbayev** is CEO and co-founder of Altbridge in New York, NY. nazym@altbridge.ai

**Andrey Kim** is CTO and co-founder of Altbridge in New York, NY. kimandr@altbridge.ai

## Abstract

Agentic AI shifts the investor's role from analytical execution to oversight. We present an agentic strategic asset allocation pipeline in which 44 specialized agents produce capital market assumptions, construct portfolios using 21 competing methods, and critique and vote on each other's outputs. A researcher agent proposes new portfolio construction methods not yet represented, and a meta agent compares past forecasts against realized returns and rewrites agent code and prompts to improve future performance. The entire pipeline is governed by the Investment Policy Statement—the same document that guides human portfolio managers can now constrain and direct autonomous agents.

# 1. Introduction

The most binding constraint in institutional asset management is not data availability or model sophistication, but the finite bandwidth of human decision-makers. A CIO can meaningfully oversee perhaps 10 to 15 investment departments, often organized along the lines of the major asset classes. A fundamental analyst can deeply cover 10 to 20 stocks. An investment committee meets monthly or quarterly, deliberates for a few hours, and ratifies decisions that must hold until the next meeting. These are organizational constraints rather than analytical or competence shortcomings. The consequence is a potential mismatch between the attention and time that human-led investment processes can supply and the complexity and speed that modern multi-asset markets demand.

Large language model (LLM) autonomous agents relax this constraint. LLMs have the characteristics of a general-purpose technology because they can be applied across many activities and can change the way work is organized (Eloundou et al. 2024). An agent in this context is not just a model that responds to a prompt; it is a goal-oriented software entity that uses tools (or runs scripts), executes multi-step reasoning, and produces structured outputs (Wang et al. 2024). Economic historians show that in past technological revolutions, the largest productivity gains occur after workflows are redesigned around new technology (David 1990, Bresnahan and Trajtenberg 1995). Early adopters of electricity replaced steam engines with electric motors while retaining belts and shafts designed for steam power. Productivity soared only after manufacturers reorganized factories around decentralized electric power. Likewise, computing initially automated existing centralized data-processing functions until personal computers distributed computing power throughout the office.

Our main contribution is conceptual. Whereas existing applications of LLM agents to investments substitute an agent for a single human task inside an otherwise unchanged workflow,[1] we redesign the strategic asset allocation (SAA) process itself as an agentic system. Agents do not just substitute for specific roles or skills but constitute the entire analytical pipeline, from capital market assumptions through portfolio construction, structured deliberation, and CIO-level combination, with the Investment Policy Statement (IPS) as the governing boundary. A single agent can execute a complete set of Capital Market Assumptions (CMAs)—for multiple asset classes using historical macro and price data, and textual and web data. Multiple agents can operate in parallel: one agent per asset class, one agent per stock, one agent per a particular portfolio construction technique. The result is an investment process that can be broader and faster than what a human team can achieve alone. Whether it is better is a different question, which we discuss in Section 5.

[1] Representative examples in investments and finance automate a single analytical role or task, like equity research agents (Zhao et al. 2025; Zhou et al. 2024), multi-agent trading systems (Xiao et al. 2024), and multi-agent financial decision-making (Yu et al. 2024).

Agents accomplish at scale what human teams lack the staff and time to implement. The agentic investment pipeline runs a specialist for each of 21 portfolio construction methods in parallel. By contrast, conventional processes generally have a single team produce CMAs for all asset classes and use a single portfolio construction method. Because every agent reasons and writes in natural language, the pipeline stages productive dissent as a repeatable protocol—structured peer review, voting, and an adversarial agent built to disagree—rather than as the disagreement a quarterly investment committee can accommodate. Our agentic architecture also allows agents to learn and acquire new skills. To our knowledge, this is the first paper to treat institutional investments as an end-to-end agentic workflow rather than to automate a single component of it.

To illustrate the agentic investment process, we run the pipeline at March 2026 across 18 liquid asset classes. The macro agent classifies a late-cycle, stagflation-tinged regime; the CMA judge marks the most expensively valued equity classes down by as much as 2.0 percentage points. The CIO agent assembles a final allocation with an expected return of 6.9% and expected volatility of 7.5%. Simulated over 1996-2026 as a static allocation, these weights would have produced a maximum drawdown of −25.6% against −34.3% for a 60/40 benchmark. We stress this is one realization of a stochastic system at a single point in time and not a backtest. Because LLMs are trained on internet-scale corpora that encode historical market data, agentic pipelines cannot be completely backtested; we mitigate look-ahead bias with proprietary techniques where we can (see Section 5). While the drawdown advantage mechanically comes from holding less equity and less risk than 60/40 in our example, what is novel is that the agentic investment architecture and its governance structure achieve this result.

## 2. Related Literature

Five capabilities distinguish agentic architectures from traditional human-based investment processes. First, **scale**: agents can run dozens, and potentially hundreds, of parallel asset-class analyses and portfolio construction methods simultaneously. Second, **productive dissent**: agents can debate, peer review, and vote on each other's proposals, embedding the productive conflict of an investment committee into a reproducible protocol. Third, **natural-language audit trails**: every agent recommendation or decision is accompanied with written justifications that can be used for ex post review and support fiduciary requirements for explainability. Fourth, **composable expertise**: adding a specialist agent—a forensic accounting module, or a new portfolio construction agent—can increase the system's capability without redesigning the pipeline. Fifth, **governable delegation**: authority can be encoded as explicit objectives, constraints, or escalation triggers, allowing institutions to increase automation while preserving human accountability over final decisions.

When multiple agents interact, new capabilities emerge. Du et al. (2023) show that multi-agent debate—where agents critique and revise each other's responses—improves both factual accuracy and reasoning quality beyond what any single agent achieves. Du et al. (2023) demonstrate that ensembles of LLM agents can match or exceed human crowd accuracy in prediction tasks. Yang et al. (2024) study how LLM agents behave under different voting rules and aggregation methods. These results motivate the structured deliberation protocol in our SAA pipeline, which combines peer review, Borda-count voting, and regime-dependent scoring to aggregate the judgments of our portfolio construction agents.[2]

Structured deliberation by agents is different from traditional forecast combination or the use of ensembles, which are well-established since Bates and Granger (1969). Combination fixes the candidate set of models and weights by a mechanical rule. In contrast, agents can add new models—as we have in a researcher agent tasked with proposing new portfolio construction methods that are not currently used. Another adversarial agent proposes a portfolio that is most orthogonal to the current set of weights, subject to certain constraints. The agents can weigh both qualitative information and quantitative information in forming their forecasts. The forecasts can also be revised before they are combined by other agents' review, which traditional combination methods cannot incorporate.

[2] On the infrastructure side, frameworks such as LangChain or LangGraph, CrewAI, AutoGen, CAMEL, MetaGPT, pi.dev, and OpenClaw provide the orchestration layer—agent-to-agent messaging, role assignment, and workflow management.

A growing body of work applies LLM agents to financial tasks. Zhao et al. (2025) deploy three specialist agents (fundamental, sentiment, and valuation) that collaborate and debate to produce buy/sell recommendations. Xiao et al. (2024) build a multi-agent trading system with distinct analyst, researcher, and trader agents. Yu et al. (2024) introduce conceptual verbal reinforcement to improve financial decision-making in a multi-agent setting, and Zhang et al. (2024) demonstrate a multimodal agent capable of processing charts, tables, and text for trading decisions. Zhou et al. (2024) present FinRobot, an orchestrator agent that chains three agents—data retrieval, financial analysis, thesis synthesis—to produce equity research reports. These architectures are not intended for SAA, do not incorporate formal voting or peer-review protocols among agents or autonomous discovery of new portfolio construction methods, and do not center governance on an IPS—features that are central to our agentic SAA pipeline.

More fundamentally, most of the research using agents in finance automates a task or role within an existing investment process, whereas we redesign the process itself—and in doing so we change the role of human decision-making. Johnson et al. (2017) identify five levels of human control in automated systems—direct, augmented, parametric, goal-oriented, and mission-capable—at each of which the human operates at a higher level of abstraction. In our setting the human moves from choosing a multiple-based or building-block model to set CMAs (direct control), to selecting which portfolio construction methods to run and their parameter bounds (parametric control), to writing the IPS that governs the entire multi-agent pipeline (goal-oriented control). The human's role is not diminished but elevated: from executing the analysis to designing and overseeing the workflow. Our architecture thus compresses work that traditionally requires teams of specialists over days or weeks into automated runs completed in minutes, while adding an agent that continuously and autonomously discovers and proposes new portfolio construction techniques—capabilities absent from the financial multi-agent systems above.

# 3. Agentic Strategic Asset Allocation

The typical SAA workflow at an institutional investor starts with the IPS, which encodes the institution's objectives, risk tolerance, time horizon, and permissible asset classes. In the next layer, investment staff or consultants then produce CMAs—expected returns, volatilities, and correlations for each asset class—drawing on macroeconomic forecasts, valuation models, and historical data. The CMAs then feed into a portfolio construction step that produces target weights, most commonly through some variant of mean-variance optimization (Markowitz 1952), with various considerations in the IPS treated as constraints. The weights are then implemented through trading, and the portfolio is monitored and periodically rebalanced.

CMAs are typically produced by a research team, or are outsourced to consultants or asset management firms, covering an asset-class universe that spans equities, fixed income, real assets, and alternatives across multiple geographies. The process to set CMAs involves loading market data, computing historical statistics, assessing the macro environment, searching for current valuations and flows, forming a forward-looking view, and writing a recommendation—a process that takes hours per asset class. The same team often runs a single portfolio construction method because running and comparing many methods is too time-consuming. The investment committee reviews one recommended portfolio, often with a sensitivity analysis. The entire cycle typically runs quarterly or semi-annually. The breadth of analysis is constrained by staffing, there is limited diversity of portfolio construction approaches, and the frequency of review is constrained by committee calendars. There is little room for structured debate among competing methodologies, systematic peer review of assumptions, or rapid adaptation to regime changes.

## *3.1 Agentic Pipeline*

The agentic architecture changes the traditional workflow into tasks assigned to one or more specialized agents, coordinated by a workflow manager.[3] Exhibit 1 illustrates the pipeline.

---

[3] See the Appendix for a description of the anatomy of an agent and sample skill files.

The IPS remains the governing document—written by humans, encoding objectives and constraints—but every subsequent stage is executed by agents operating within IPS bounds.

The pipeline proceeds in six steps:

1. A **macro agent** classifies the current economic regime (expansion, late-cycle, recession, or recovery) based on macro and market indicators, providing a regime signal that conditions all downstream analysis.
2. **Asset class (AC) agents** run in parallel, each producing CMAs—an expected return, a volatility estimate, and a confidence level—along with a written investment-case memo for its assigned asset class.
3. A **covariance agent** estimates the asset class covariance matrix using historical data and macro forecasts.
4. **Portfolio construction (PC) agents** take the CMAs from step (2) and the covariance matrix from step (3) to independently construct a proposed portfolio, using methods ranging from equal-weight and inverse-volatility heuristics to mean-variance optimization, risk parity, hierarchical risk parity, and Total Portfolio Approach methods.
5. A **multi-agent strategy review** subjects all PC proposals to peer review, risk assessment, and voting.
6. A **CIO agent** scores, combines, and selects from the surviving proposals using an ensemble of seven combination methods. The CIO agent produces a board memo summarizing the recommendation, the reasoning, and the dissenting views.

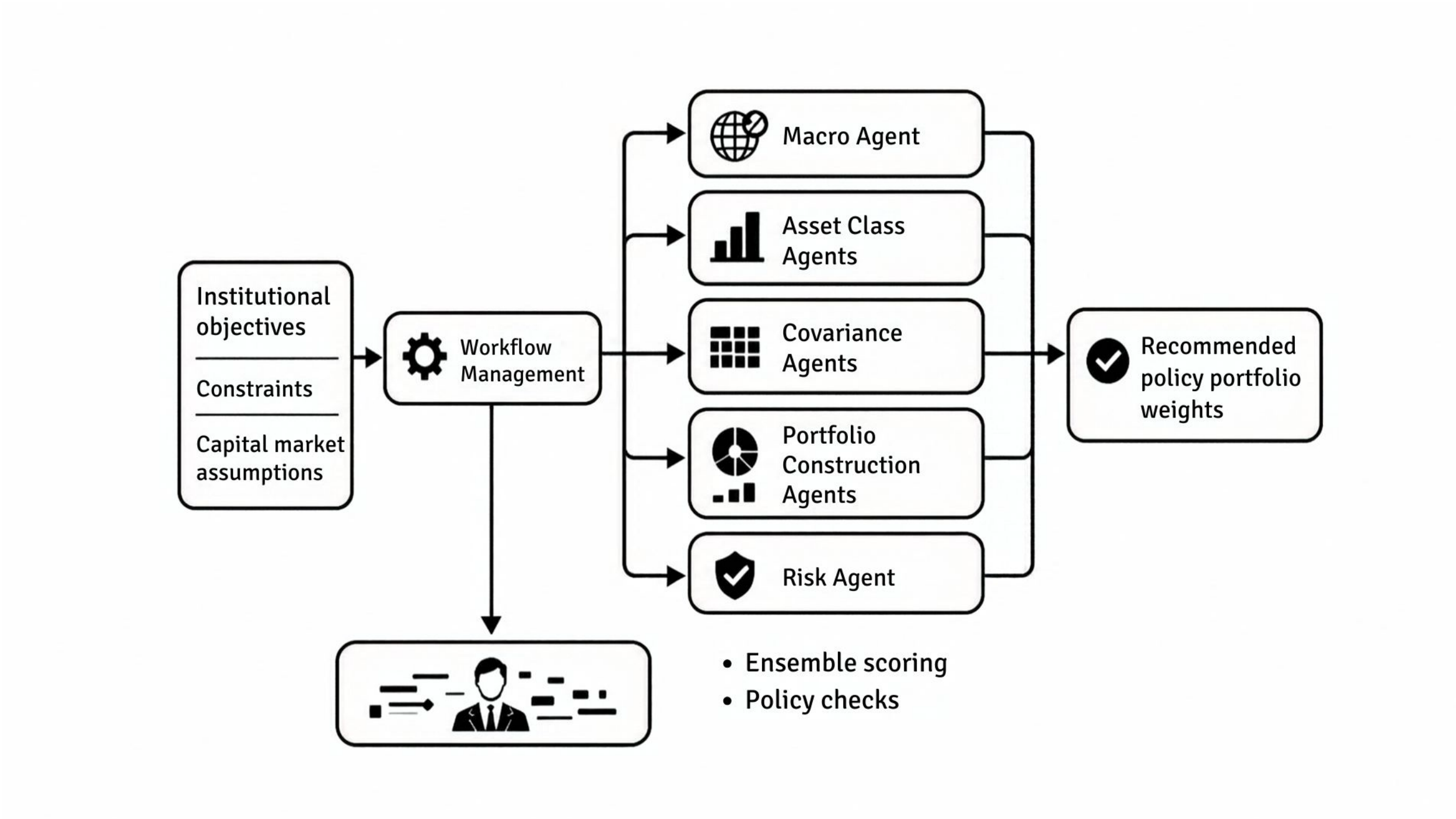

**Exhibit 1**: Agentic Strategic Asset Allocation Architecture

What distinguishes the agentic SAA workflow from a conventional quantitative SAA workflow is not just parallelism—running two dozen optimizers simultaneously is straightforward—but the fact that each agent reasons in natural language, produces a written justification for its output, and can read and respond to the outputs of other agents. The strategy review, described below in Section 3.4, is a structured deliberation in which agents critique, vote on, and revise each other's proposals. No traditional optimizer framework does this.

To illustrate, consider the macro agent, which runs first in the pipeline (see Exhibit 2). The macro agent first fetches macro and market data. The agent also searches the web for real-time readings of both numeric and textual information. With the data, it scores four dimensions—growth, inflation, monetary policy, and financial conditions—using a weighted scoring framework to classify the current regime with a confidence level into one of four regimes: expansion, late-cycle, recession, or recovery. Finally, the agent writes a file with the macro regimes and a narrative report. The output is consumed by all downstream asset class agents.

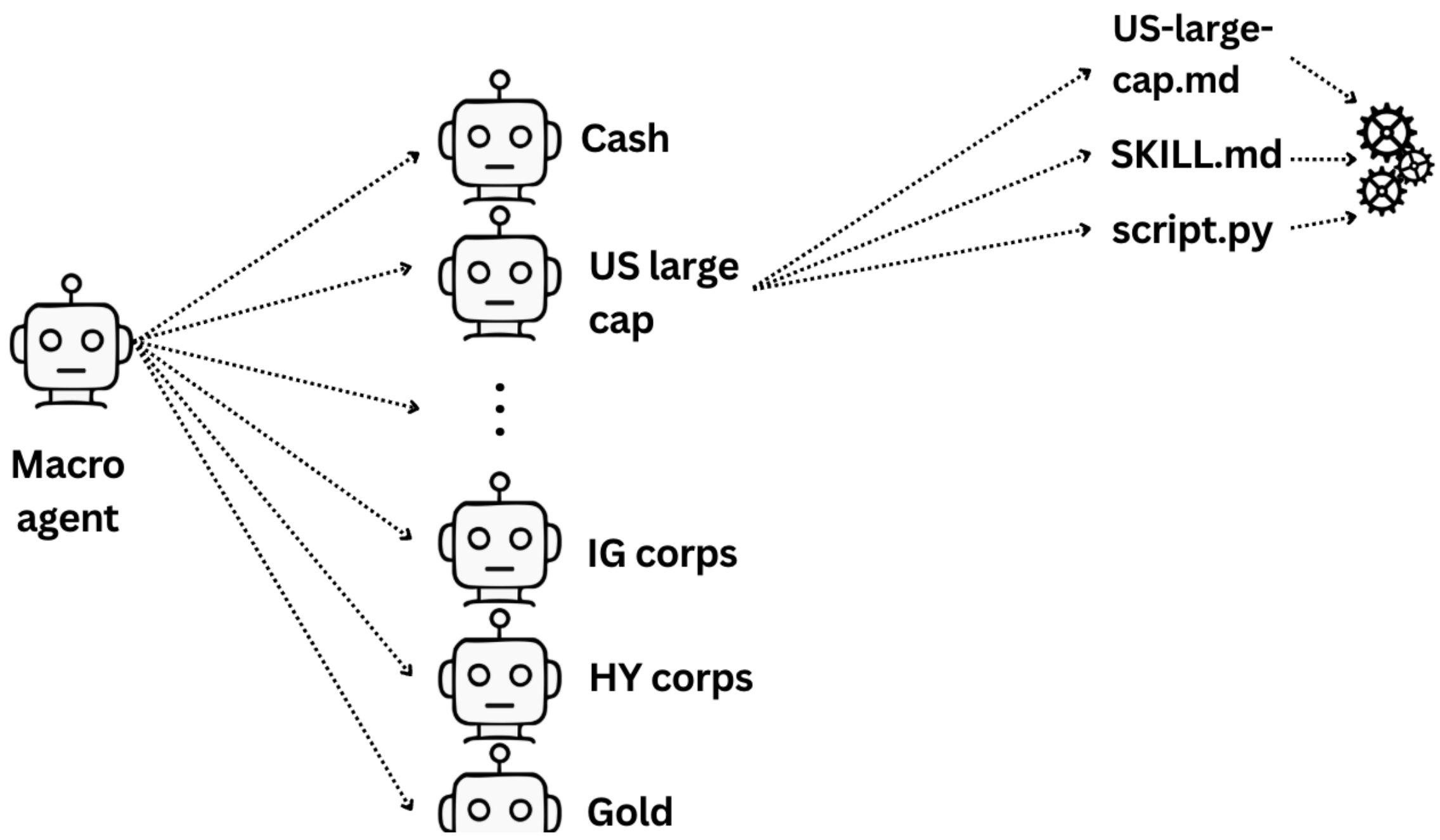

**Exhibit 2**: Asset Class (AC) Agents

## 3.2 Asset Class (AC) Agents

Each AC agent takes the macro view produced by the macro agent, and develops asset class-specific views. Exhibit 2 illustrates the workflow. A feature of the AC agents is that they evaluate CMAs using several different methods and select one, or combine the methods, to produce the final set of expected returns.

As an example, the US Large Cap agent analyzes the S&P 500 and, after loading the macro view, runs historical, valuation, technical, sentiment, and scenario analysis. The agent computes expected returns from several methods: the historical equity risk premium (ERP) added to the current risk-free rate; a regime-adjusted ERP conditioned on the current macro cycle; implied returns consistent with Black–Litterman (1992) equilibrium weights; an inverse Gordon (1959) building-block model following Grinold and Kroner (2002) that sums dividend yield, earnings growth, and valuation change; the implied ERP from the Campbell and Shiller (1998) CAPE; and consensus survey forecasts. Each method returns a point estimate, a confidence score, a component breakdown, and a rationale. As another example, the US Small Cap agent's key considerations include taking into account comparisons relative to large cap equities and corporate debt. The corporate bond agents emphasize credit-spread duration and sector concentration in forming their CMA estimates.

The judgment step follows. The AC agents apply a structured LLM-as-judge framework (Zheng et al. 2023) to select or combine the final estimate. The agents read all candidate CMA estimates alongside the macro regime, asset-level signals, and historical statistics, and then evaluate which methods' assumptions best fit the current environment. The agents favor valuation-based methods when valuations are stretched, tilt toward regime-conditional estimates in late-cycle or recessionary periods, and default to the auto-blend when methods broadly agree—subject to the hard constraint that the final estimate must lie within the range of the candidate methods.

## 3.3 Portfolio Construction (PC) Agents

The CMAs and covariance matrix feed into 21 PC agents, each implementing a distinct optimization objective. The first 19 PC agents run in parallel and implement the fixed portfolio construction methods listed in Exhibit 3. The 20th agent, the PC researcher, searches for a new portfolio construction method currently not spanned by the fixed list, and runs at the same time. The 21st agent, the adversarial diversifier, maximizes tracking variance relative to the ensemble centroid of the other agents, and executes after the initial 20 have finished. The agents are organized into five categories, with four to six methods per category (except the new agentic category), and are summarized in Exhibit 3.[4]

The first category of PC agents is *heuristic* methods, which avoid optimization-driven estimation error and dominate when expected returns are poorly measured (DeMiguel, Garlappi, and Uppal 2009). The second category is *return-optimized* methods which explicitly use return forecasts from the asset class agents. The *risk-structured* methods optimize risk metrics without explicit return forecasts, on the premise that the covariance matrix is more reliably estimated than expected returns (Merton 1980). *Non-traditional* methods address limitations of variance-based frameworks by optimizing expected shortfall, drawdown profiles, or multi-factor risk budgets. Among the methods in this category that is gaining popularity among institutional investors is Total Portfolio Approach, which is not bound by traditional asset classes (Ang, Brandt, and Denison 2014; Gilmore and Simonian 2025).

[4] The PC methods are initially a design choice, but as the agents discover and evaluate new PC methods, and decommission other PC approaches, the size and type of the PC registry will become endogenous. All of the PC methods are standard reference implementations that an institutional investor would recognize. We give sample references which contain more details on each PC method in Exhibit 3.

The agentic category contains methods unique to agents: a *Researcher* agent and an *Adversarial Diversifier* agent. Rather than implementing a fixed optimization objective, the PC researcher agent explores the literature on portfolio construction methods, identifies objectives not yet represented in the pipeline, and proposes a novel method not spanned by the current registry of PC methods. In the March 2026 run, the PC researcher proposes a maximum entropy portfolio (Bera and Park 2008) that maximizes the Shannon entropy of portfolio weights subject to a minimum Sharpe ratio floor. It is expected that successful new PC methods will be added to the registry of PC agents, and a portfolio review process can also cull unsuccessful PC methods.

The PC adversarial diversifier agent operates as follows. It maximizes tracking variance relative to the ensemble centroid (the mean of all other PC weights), subject to a Sharpe ratio floor of 75% of the maximum Sharpe portfolio. Thus, it constructs the most orthogonal portfolio to the other 20 PC agents (including the PC researcher agent) subject to the Sharpe ratio floor. The adversarial diversifier is not intended as a standalone recommendation; it surfaces allocation ideas that are overlooked by conventional methods and can reduce pairwise weight overlaps when incorporated into the meta-portfolio.

## *3.4 Portfolio Construction (PC) Strategy Review*

The PC strategy review is a process in which the 21 PC agents critique and rank each other's work, similar to the multi-agent debate and voting mechanisms of Du et al. (2023) but in a setting where the agents hold diverse priors encoded by their PC category. The review proceeds in stages, summarized in Exhibit 4. First, a Chief Risk Officer (CRO) agent produces a standardized risk report for each candidate portfolio, which covers standard risk metrics like ex-ante and backtest volatility, value-at-risk, maximum drawdown, concentration metrics, factor tilts, and IPS compliance. The CRO agent is a neutral assessor: it scores risk and produces commentary, but does not vote.

| Category | Method | Sample Reference |
|---|---|---|
| **A. Heuristic** | Equal weight (1/N) | DeMiguel, Garlappi, and Uppal (2009) |
| | Market-cap weight | Sharpe (1964) |
| | Inverse volatility | Kirby and Ostdiek (2012) |
| | Inverse variance | Kirby and Ostdiek (2012) |
| | Volatility targeting | Moreira and Muir (2017) |
| **B. Return-optimized** | Maximum Sharpe ratio | Markowitz (1952) |
| | Black–Litterman | Black and Litterman (1992) |
| | Robust mean-variance | Goldfarb and Iyengar (2003) |
| | Resampled efficient frontier | Michaud (1998) |
| | Mean–downside risk | Sortino and van der Meer (1991) |
| **C. Risk-structured** | Global minimum variance | Clarke, de Silva, and Thorley (2006) |
| | Risk parity (Equal risk contribution) | Maillard, Roncalli, and Teïletche (2010) |
| | Hierarchical risk parity | López de Prado (2016) |
| | Maximum diversification | Choueifaty and Coignard (2008) |
| | Minimum correlation | Varadi et al. (2012) |
| **D. Non-traditional** | CVaR optimization | Rockafellar and Uryasev (2000) |
| | Max drawdown-constrained | Chekhlov, Uryasev, and Zabarankin (2005) |
| | Tail-risk parity | Boudt, Carl, and Peterson (2013) |
| | Total Portfolio Approach (TPA) two-factor (equity and bonds) | Ang, Brandt, and Denison (2014) |
| **E. Agentic** | Adversarial diversifier | — |
| | Researcher | — |

**Exhibit 3**: Portfolio Construction (PC) Agents

Each PC agent reviews two peers—one from its own category (intra-category review, which is more likely to identify technical errors within a shared framework) and one from a different category (inter-category review, which can challenge foundational assumptions from a contrasting worldview).[5] Assignments are randomized with a recorded seed, producing 42 reviews across the 21 candidates. All reviews are released simultaneously so that every agent can read every review before voting.

[5] The two-reviewer design keeps the cost of the agent reviews linear. If every agent referees every other agent, the computation cost is quadratic in the number of agents.

Voting follows a modified Borda count. Each agent submits a top-five ranking (awarding 5, 4, 3, 2, and 1 points) and a bottom flag (−2 points), excluding itself. This structure surfaces both consensus and dissent simultaneously: a method that receives near-unanimous top-five placement reflects broad agreement, while a method flagged bottom-one by a majority signals a clear rejection. The vote totals are then blended with a quantitative metric score (a weighted composite of backtest Sharpe, IPS compliance, diversification, regime fit, estimation risk, and CMA utilization) using a regime-dependent weight. This blending ensures that the final ranking reflects both collective peer judgment and objective performance metrics. A diversity constraint requires the top-five shortlist to include representation from at least three of the five families, preventing a single philosophical school from monopolizing the recommendation set.

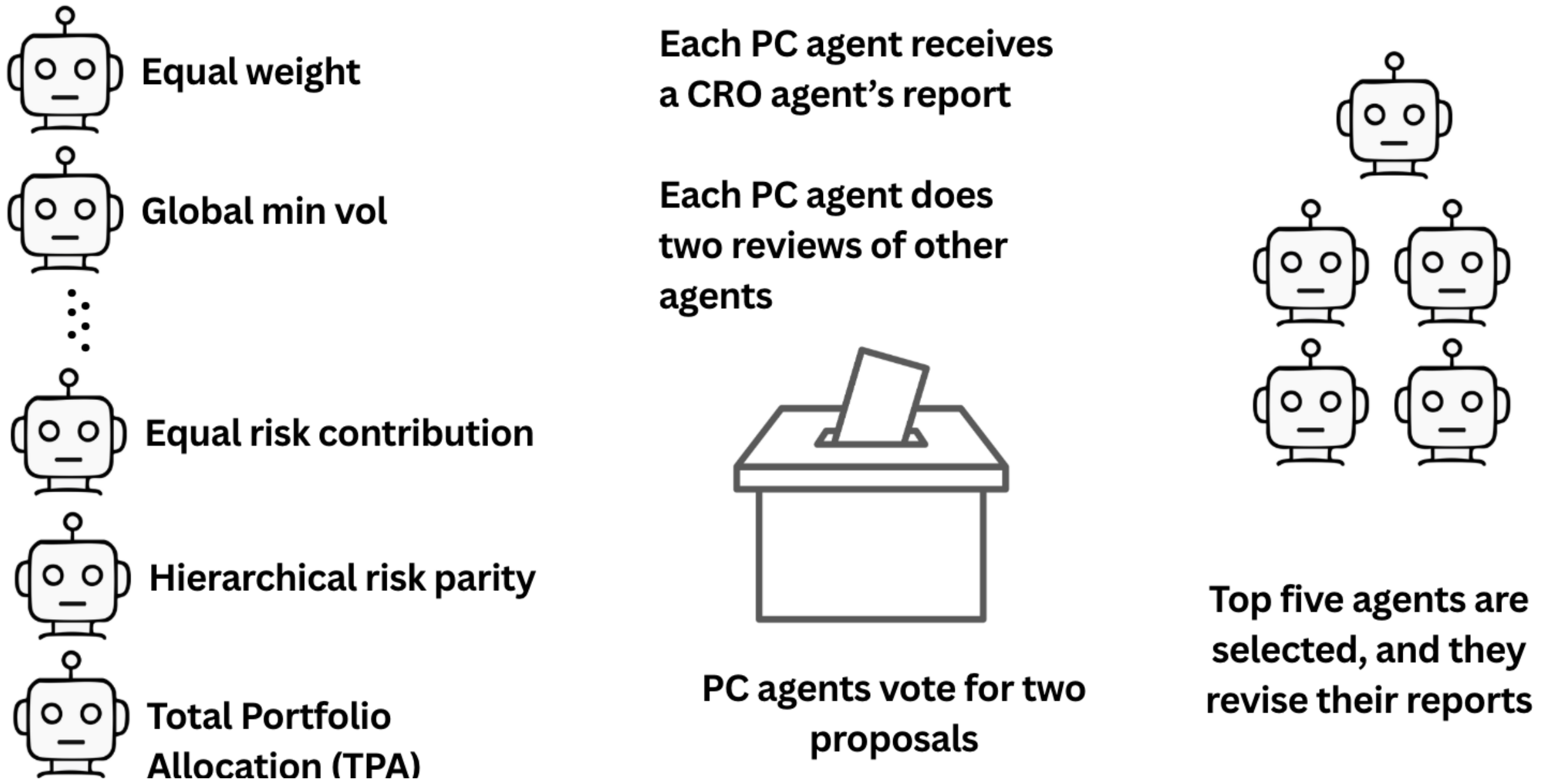

**Exhibit 4**: Portfolio Construction (PC) Strategy Review

Finally, the top-five ranking PC agents then revise their proposals, taking into account the peer reviews and the CRO agent's report. The revisions mostly take into account the two specific reviews, but since the agents have full information, they can revise their structured output to incorporate comments from any PC agent.

## *3.5 CIO Agent*

The CIO agent receives the strategy review output—the 21 candidate portfolios, their peer reviews, CRO risk reports, vote tallies, and metric scores—and constructs the final recommended allocation operating as an LLM-as-judge. (See Exhibit 5.) In addition to being able to choose a given portfolio method, the CIO agent has access to several ensemble techniques: simple average, inverse tracking-error weighting, backtest Sharpe-ratio weighting, meta-optimization that treats PC portfolios as "assets" in a second-level optimization, regime-conditional weighting (which varies method weights by macro regime), composite score weighting, and a trimmed mean. The CIO agent evaluates each ensemble on the same diagnostic suite used for individual portfolios—performance statistics and checking IPS compliance (which is non-negotiable)—and selects the ensemble method best suited to the current regime with a written rationale.

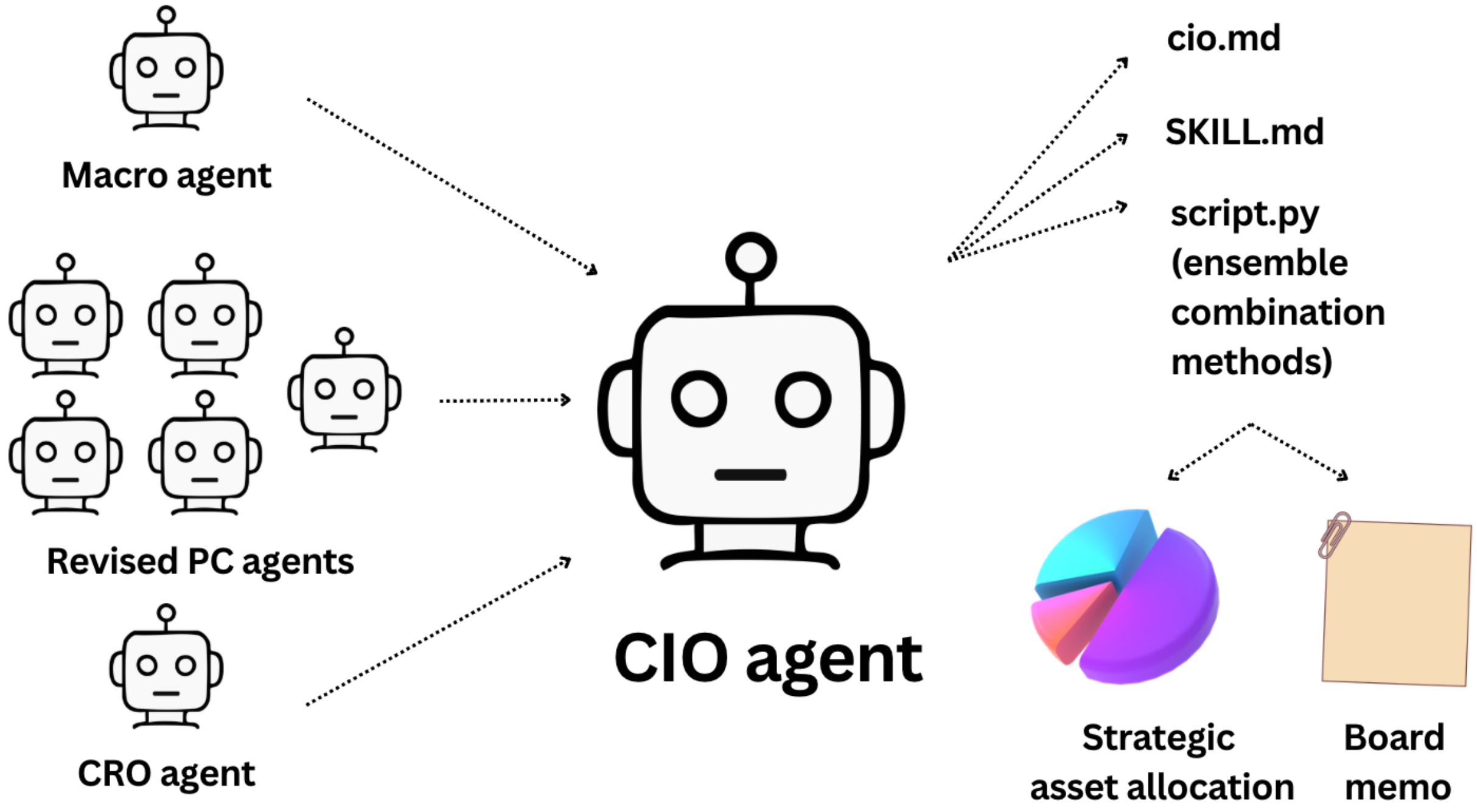

**Exhibit 5**: CIO Agent

A crucial role of the CIO agent is not just to produce weights—the CIO agent explains *why* the portfolio was chosen, what assumptions drive it, and what would cause the portfolio to no longer be valid. The CIO agent also produces a board memo, which summarizes the recommendation in a governance document written for non-technical stakeholders. The memo covers the recommended allocation and its expected performance relative to a 60/40 equity-bond benchmark, the macro rationale, the largest positions, changes since the last review, key risks to monitor together with a quarterly rebalancing plan with off-cycle drift triggers, and an IPS compliance statement. The board memo thus closes the loop between the autonomous agentic pipeline and institutional oversight: the agents execute every analytical step, but the output is a document that a board of trustees or investment committee can read, challenge, and approve—preserving the governance framework that Johnson et al. (2017) advocate at every level of control abstraction.

## 4. Illustrative Run

We illustrate the pipeline with a single run in March 2026. There are two caveats. First, this is one realization of a stochastic system at a single point in time; a run for the same historical date (March 2026) produced at a different time (say September 2026) may produce different numbers. Thus, we present the illustrative run as a worked example of how agents can interact and produce an investment decision, not as evidence of outperformance. Second, agentic pipelines cannot be cleanly backtested. Reproducibility is a recognized challenge for LLM-based systems: identical prompts under nominally deterministic settings (temperature zero, fixed seed) can still yield different outputs on different batch sizes and hardware (Atil et al. 2025), and model behavior can drift as providers update models (Chen, Zaharia, and Zou 2023). LLMs inevitably encode historical financial data, so any historical "out-of-sample" evaluation is contaminated by look-ahead bias.[6] The figures below should be read as an

[6] We mitigate this with proprietary techniques. See also Section 5.1 for additional discussion. We pin the model version and sampling parameters within a review cycle, keep deterministic computation in code rather than

illustration of how the agents reason and govern in an investment workflow, not as a return forecast or a track record. We comment on these points further in Section 5.

## 4.1 *Investment Policy Statement (IPS)*

The most important part of the agentic SAA framework is the Investment Policy Statement (IPS). In our example, the IPS has three layers. The first layer specifies the *investment universe* of 18 liquid asset classes, each investable through exchange-traded funds. The six equity classes are US Large Cap, US Small Cap, US Value, US Growth, International Developed, and Emerging Markets. The eight fixed income classes are Short-Term Treasuries, Intermediate Treasuries, Long-Term Treasuries, Investment-Grade Corporates, High-Yield Corporates, International Sovereign Bonds, International Corporates, and USD Emerging Market Debt. The four remaining classes are REITs, Gold, Commodities, and Cash. The second layer specifies the *objective*: a target real return of CPI +3.0–4.0%, an expected volatility band of 8–12%, and a maximum drawdown limit of −25% peak-to-trough. The third layer is the *active risk budget*: the portfolio's ex-ante tracking error relative to the 60/40 equity-bond benchmark must not exceed 6%.

One governance point is worth emphasizing because it bears on how the results should be interpreted. We treat the IPS bounds as a mix of hard constraints and soft targets. The asset universe and the 6% tracking error constraint are hard constraints: the CRO agent enforces them and the CIO agent may not breach them. The return target, the 8−12% volatility band, and the −25% drawdown limit are soft targets that the agents seek to satisfy but may miss when doing so better serves the IPS's overriding objective—in which case the deviation is flagged in the board memo for the committee rather than silently overridden. (This is the case that we find in the illustrative run, as we discuss below.) This distinction matters in the run below: the recommended portfolio's ex-ante volatility of 7.54% sits

---

use the LLM, and homogenize and obfuscate agent inputs so that judgments are anchored to the disclosed evidence rather than to LLMs' knowledge of a specific security or period. The version and type of each frontier foundation model used by each agent is documented, along with temperature, top-k, top-p, and other parameters controlling the output of tokens, peer-review pairing seeds, full agent description files, skill files, output contracts, prompts, and natural language reasoning and audit trails. The Appendix contains some examples of these files. All numerical computation, statistics, and optimizer runs are executed in code, not in the LLM, and are fully replicable.

modestly below the lower edge of the 8–12% band, a deliberate choice in a late-cycle regime that the CRO flags and the CIO accepts; the CIO does not "correct" this undershooting of risk by substituting riskier assets. We believe that treating these bounds as guardrails, with explanation required, rather than as inviolable limits, is closer to how investment committees actually use an IPS, and it incorporates a natural-language audit trail that a human can use to approve or reject each deviation.

## 4.2 Macro Regime

The macro agent classifies the current environment as *late-cycle with stagflationary risk* at medium-high confidence. US real GDP growth has decelerated in the most recent revision, while nonfarm payrolls turn negative in February. Inflation is moderating toward target (CPI 2.4%, core 2.5%) but is expected to re-accelerate toward 3% following an oil supply shock from the Iranian conflict that is pushing up the price of Brent crude. The macro agent estimates a baseline recession probability of 25–35%, consistent with estimates from major forecasters. This regime classification propagates to all downstream agents.

## 4.3 Asset Class (AC) Agents

We concentrate on the equity AC agents to illustrate the effect of LLM-as-judge. Exhibit 6 reports the method-level CMA estimates, the confidence-weighted auto-blend, and the CMA judge's final selection for the six equity asset classes. Each agent defines six expected-return methods producing candidate estimates that span a wide range. The judge reads all candidates alongside the macro regime, asset-level signals, and historical statistics, and constructs a custom blend with explicit weights and a written rationale.

A feature of Exhibit 6 is the cross-sectional pattern of the LLM-as-judge's adjustments. All the AC agents acting as judges choose custom blends rather than a single method, but the magnitude of adjustment varies systematically with valuation. US Growth—the most expensive asset class (CAPE ≈ 31, earnings yield 3.2% versus a 3.7% risk-free rate)—receives the largest markdown of −2.0% to the auto-blend, as the judge assigns 60% weight to valuation-based methods and only 5% to the historical equity risk premium (ERP).

| Asset Class | Hist. ERP | Regime Adj. | BL Equil. | Inv. Gordon | Impl. ERP | Auto-Blend | Judge | Δ Judge vs. Auto Blend |
|---|---|---|---|---|---|---|---|---|
| US Large Cap | 12.5 | 9.8 | 9.3 | 4.3 | 4.0 | 7.9 | **6.8** | −1.1 |
| US Growth | 13.3 | 10.0 | 9.7 | 3.5 | 3.2 | 8.2 | **6.2** | −2.0 |
| US Value | 11.7 | 9.5 | 9.0 | 4.8 | 4.9 | 8.0 | **7.5** | −0.5 |
| US Small Cap | 12.1 | 9.8 | 10.2 | 4.1 | 5.6 | 8.2 | **8.2** | 0.0 |
| Intl Developed | 7.7 | 6.2 | 9.3 | 6.6 | 5.7 | 7.0 | **6.7** | −0.3 |
| Emerging Markets | 10.9 | 9.2 | 10.4 | 5.3 | 6.9 | 8.4 | **8.2** | −0.2 |
| REITs | 12.1 | 11.2 | 8.7 | 7.1 | 3.4 | 8.5 | **8.3** | −0.2 |

**Exhibit 6**: CMA Method Estimates and Judge Selections for Equity Asset Classes and REITs (%, nominal, 3-year horizon)

US Large Cap (CAPE ≈ 25) is marked down by −1.1%. At the other end, Emerging Markets (CAPE ≈ 13) and REITs receive adjustments of only −0.2%, while US Small Cap is essentially unchanged. The agents are not uniformly bearish; they are skeptical of backward-looking return estimates for asset classes where current valuations are most stretched. This valuation-conditioned skepticism is an economically interpretable behavior that emerges from the judgment protocol rather than being hard-coded.[7]

## 4.4 Portfolio Construction (PC) Strategy Review

The 21 PC agents—the 19 canonical methods listed in Exhibit 3 plus the max-entropy method proposed by the PC researcher agent and the adversarial diversifier—each referee two other agents' proposals and vote. The voting results are summarized in Exhibit 7. The composite score is computed by 40% of the normalized vote score and 60% of the normalized performance metrics. The Borda-count vote produces a clear hierarchy: maximum diversification ranks first, followed by Black–Litterman and risk parity methods. The top five are rounded out by risk parity strategies: regular risk parity and hierarchical risk parity. At the bottom are the CVaR optimization and the adversarial diversifier, which trigger dissent reports. These results are consistent with a late-cycle regime with relatively large uncertainty on the expected returns; not surprisingly the top methods that rely primarily on the covariance structure and not the CMAs (Categories C and D) are favored by the agents.

[7] Another interpretation is that the agents are biased to a bearish outlook (but not for US Small Cap which is less expensive than other equities at this time). Separating the two interpretations requires running the judgment step in periods when equities were cheap like March 2020. We cannot cleanly run true out-of-sample experiments with agentic models because they are trained through the present, but we do not observe permanent bearish tilts in runs of the agentic investment process over past data.

| Rank | Method | Category | Vote | Metric | Composite |
|---|---|---|---|---|---|
| 1 | Maximum Diversification | C: Risk-Structured | 96 | 0.553 | 1.000 |
| 2 | Black-Litterman | B: Return-Optimized | 76 | 0.551 | 0.936 |
| 3 | Risk Parity | C: Risk-Structured | 42 | 0.503 | 0.750 |
| 4 | Hierarchical Risk Parity | C: Risk-Structured | 41 | 0.497 | 0.737 |
| 5 | Tail Risk Parity | D: Non-Traditional | 22 | 0.510 | 0.702 |
| 6 | Total Portfolio Approach | D: Non-Traditional | 22 | 0.479 | 0.648 |
| 7 | Resampled Efficient Frontier | B: Return-Optimized | 5 | 0.468 | 0.578 |
| 8 | Equal Weight | A: Heuristic | 3 | 0.468 | 0.572 |
| 9 | Mean-Downside Risk (Sortino) | B: Return-Optimized | 1 | 0.459 | 0.549 |
| 10 | Maximum Sharpe Ratio | B: Return-Optimized | 4 | 0.438 | 0.522 |
| 11 | Maximum Entropy | E: Agentic | 3 | 0.406 | 0.464 |
| 12 | Inverse Volatility | A: Heuristic | 0 | 0.407 | 0.457 |
| 13 | Robust Mean-Variance | B: Return-Optimized | 0 | 0.406 | 0.456 |
| 14 | Inverse Variance | A: Heuristic | 0 | 0.365 | 0.386 |
| 15 | Max Drawdown Constrained | D: Non-Traditional | 0 | 0.362 | 0.380 |
| 16 | Global Minimum Variance | C: Risk-Structured | 0 | 0.359 | 0.375 |
| 17 | Minimum Correlation | C: Risk-Structured | 0 | 0.355 | 0.368 |
| 18 | Volatility Targeting | A: Heuristic | 0 | 0.304 | 0.280 |
| 19 | Market-Cap Weight | A: Heuristic | 0 | 0.253 | 0.191 |
| 20 | CVaR Minimization | D: Non-Traditional | -6 | 0.239 | 0.150 |
| 21 | Adversarial Diversifier | E: Agentic | -36 | 0.205 | 0.000 |

**Exhibit 7**: Agent Voting on PC Methods

As described in Section 3.3, the PC researcher agent identifies a new PC technique not currently spanned by the candidate PC agents—maximum entropy. The maximum entropy portfolio delivers an historical Sharpe ratio of 0.41, which is between the return-optimized and risk-structured categories. In the strategy review vote, the max-entropy finishes 11th, a middling rank that reflects its novelty rather than poor quality. Only the adversarial diversifier votes it into a top-5, while most other agents rank it lower because they lack prior experience of the method. Based on the agents' voting, a production pipeline would likely add this method to the roster of PC agents in subsequent runs.

The adversarial diversifier is expected to be rejected by its peers because it is, by construction, orthogonal—and that rejection is a feature, not a flaw. In the strategy review, 18 agents voted it at the bottom. But the adversarial diversifier is valuable at the CIO ensemble stage, to which we now turn.

## 4.5 CIO Agent

The CIO agent, operating as an LLM-as-judge, scores each of the 21 PC methods on six dimensions—backtest Sharpe (25%), IPS compliance (15%), diversification (15%), regime fit (20%), estimation robustness (15%), and CMA utilization (10%)—and constructs the final portfolio from an ensemble. In the March 2026 run, the CIO agent selects the inverse-tracking-error-weighted ensemble, which assigns higher weight to methods whose portfolios are closer to the ensemble centroid. The resulting portfolio has an expected return of 6.87%, volatility of 7.54%, a backtested Sharpe ratio of 0.43, and an effective number of assets of 11.2 (Meucci 2009).[8] The ex-ante tracking error versus a 60/40 benchmark is 2.41%. We also note that the recommended portfolio's ex-ante volatility of 7.54% sits below the IPS's recommended 8-12% band, but this is a soft constraint. This is exactly the kind of deviation that is captured in the board memo that goes to the investment committee for discussion and ratification.

The top ensemble weights are reported in Exhibit 8 and are the market-cap weight (11.1%), volatility targeting (6.7%), equal weight (6.0%), inverse variance (6.0%), and max-entropy (5.6%). It is notable that the max-entropy method is "discovered" by the PC researcher agent, and the CIO agent assigns a non-zero weight to the adversarial diversifier even though its standalone metrics (effective number of assets of 2.4, maximum drawdown of −46.3%) would disqualify it under any single-method selection rule, because the scoring gives the LLM-as-judge CIO agent latitude to value ensemble diversification alongside other criteria.

[8] For comparison, the 18-asset universe has an effective number of assets equal to 18 by construction and the 60/40 benchmark has an effective number of assets of approximately 2.

| Method | Category | Ensemble Weight | Vote Rank | Composite |
|---|---|---|---|---|
| Market-Cap Weight | Heuristic | 11.1% | 19 | 0.191 |
| Volatility Targeting | Heuristic | 6.7% | 18 | 0.280 |
| Equal Weight | Heuristic | 6.0% | 8 | 0.572 |
| Inverse Variance | Heuristic | 6.0% | 14 | 0.386 |
| Maximum Entropy | Agentic (PC-Researcher) | 5.6% | 11 | 0.464 |
| Inverse Volatility | Heuristic | 5.6% | 12 | 0.457 |
| Hierarchical Risk Parity | Risk-Structured | 5.3% | 4 | 0.737 |
| Tail Risk Parity | Non-Traditional | 5.3% | 5 | 0.702 |
| Risk Parity | Risk-Structured | 5.1% | 3 | 0.750 |
| CVaR Minimization | Non-Traditional | 4.2% | 20 | 0.150 |
| Robust Mean-Variance | Return-Optimized | 4.1% | 13 | 0.456 |
| Resampled Efficient Frontier | Return-Optimized | 4.1% | 7 | 0.578 |
| Global Minimum Variance | Risk-Structured | 4.0% | 16 | 0.375 |
| Maximum Sharpe Ratio | Return-Optimized | 4.0% | 10 | 0.522 |
| Mean-Downside Risk (Sortino) | Return-Optimized | 4.0% | 9 | 0.549 |
| Max Drawdown Constrained | Non-Traditional | 3.6% | 15 | 0.380 |
| Minimum Correlation | Risk-Structured | 3.5% | 17 | 0.368 |
| Black-Litterman | Return-Optimized | 3.3% | 2 | 0.936 |
| Maximum Diversification | Risk-Structured | 3.1% | 1 | 1.000 |
| Adversarial Diversifier | Agentic | 2.7% | 21 | 0.000 |
| Total Portfolio Approach | Non-Traditional | 2.7% | 6 | 0.648 |

**Exhibit 8**: Ensemble Weights Across Portfolio Construction Methods

The adversarial diversifier contributes allocations that are, by construction, orthogonal to consensus. This reduces pairwise weight overlaps across the ensemble inputs, expanding the effective spanning set of portfolios available to the CIO. The mechanism is analogous to boosting in machine learning: each weak learner is individually suboptimal, but the ensemble benefits precisely from learners that make forecasting errors in uncorrelated dimensions. Thus, the adversarial diversifier provides the CIO agent with exposure to regions of the allocation space that no PC agent peer-approved method would have discovered.

A natural question is why the CIO's ensemble weights (Exhibit 8) do not track the deliberation ranking after the agents' voting process (Exhibit 7): the CIO agent assigns its largest weight to market-cap weighting (ranked 19th by the agents in Exhibit 7) and only 3.1% to maximum diversification (ranked first by the agents). This divergence is intentional, and we regard it as a feature of the architecture rather than a defect. The peer-review, CRO, and Borda-vote layer is a screening, revision, and dissent-surfacing mechanism: it validates the candidate set, forces the top five to revise, removes methods that fail the IPS or contain technical errors, and gives the CIO agent structured information—including why the adversarial diversifier is an outlier. The CIO agent's ensemble is a separate combination decision; it is not a weight-setting rule. At March 2026, the CIO agent's weighting rewards centrality—proximity to the consensus of all 21 methods—rather than single-method quality. In a late-cycle regime where expected returns are poorly measured, concentrating in the single highest-ranked method is what the CIO agent wishes to avoid, so a robust choice is the region of the allocation space where many methods agree. Maximum diversification ranks first because peers admire its risk properties, but it is an extreme, near-orthogonal portfolio that sits far from the centroid and is therefore down-weighted. In other words, the Borda rank measures peer respect, whereas the CIO agent's weighting at March 2026 emphasizes robustness; the two diverge.

The final allocation is reported in Exhibit 9. Relative to a 60/40 benchmark, the target portfolio is modestly underweight equity (44.9% versus 60%), roughly in line on fixed income (41.7% versus 40%), and carries an 8.1% cash position alongside a 5.1% real assets sleeve. Within equity, the allocation exhibits an international tilt and within fixed income, the portfolio concentrates in duration with Intermediate and Long-Term Treasuries together accounting for 23.1% of total weight. Credit-spread sectors (High Yield, IG Corporates, EM Debt, and International Corporates) are held in modest satellite positions totaling 6.5%.

| Asset Class | Category | Weight | Risk Contrib. |
|---|---|---|---|
| Intl Developed | Equity | 15.9% | 29.8% |
| Intermediate Treasuries | Fixed Income | 14.7% | 2.9% |
| US Large Cap | Equity | 8.9% | 15.3% |
| Long-Term Treasuries | Fixed Income | 8.4% | 2.9% |
| Cash | Cash | 8.1% | 0.1% |
| Short-Term Treasuries | Fixed Income | 7.3% | 0.2% |
| US Value | Equity | 7.2% | 11.9% |
| Emerging Markets | Equity | 4.9% | 10.9% |
| Intl Sovereigns | Fixed Income | 4.8% | 0.7% |
| US Growth | Equity | 4.4% | 8.2% |
| US Small Cap | Equity | 3.6% | 7.2% |
| IG Corporates | Fixed Income | 2.4% | 1.2% |
| Gold | Real Assets | 2.1% | 1.1% |
| HY Corporates | Fixed Income | 1.6% | 1.3% |
| Commodities | Real Assets | 1.6% | 1.3% |
| REITs | Real Assets | 1.4% | 2.3% |
| Intl Corporates | Fixed Income | 1.3% | 1.3% |
| USD EM Debt | Fixed Income | 1.2% | 1.2% |

**Exhibit 9**: Final Portfolio Weights

We stress that this comparison is illustrative, and is certainly not evidence of outperformance. The agentic pipeline cannot be cleanly backtested, and the more favorable drawdown is largely mechanical: the recommended portfolio holds 44.9% equity versus 60% for the benchmark, with an 8.1% cash sleeve and an ex-ante volatility (7.54%) well below that of 60/40, so a smaller drawdown is what one would expect from a lower-risk portfolio. We also note that the realized drawdown of −25.6% sits just beyond the IPS's −25% soft target, which is the type of deviation the board memo would discuss. We emphasize the result is not performance but process: the allocation is the documented, auditable output of a governed agentic deliberation.

We now discuss the shortcomings of this illustrative agentic investment example.

# 5. Limitations and Governance

There are several risks in our agentic SAA framework, some of which are different from the risks in a traditional human-based process.

## 5.1 Data Leakage

A cornerstone of quantitative investment research is the backtest—using LLMs, unfortunately, introduces new forms of information leakage, or look-ahead bias. Cheng, Chang, and Wu (2025) document that LLMs trained on internet-scale corpora inevitably contain past information that contaminates point-in-time forecasts. This means that LLMs encode historical financial data and market outcomes that are unavailable at the historical decision date.[9] The most rigorous solution is to use LLMs that have been trained only up to a point in time. Yan et al. (2026) and Kelly et al. (2026) construct models with annual or monthly cutoffs but these models are much smaller than the leading frontier LLMs and are computationally intensive to train. Point-in-time frontier proprietary models are generally not available.

Agents call quantitative methods—like return forecasting methods and optimization techniques—that are estimated and evaluated with standard out-of-sample backtests. The new distinct risk of information leakage arises at the agentic layer: LLM-as-judge, interpretation, outputs, and agent-to-agent interaction. This is a different form of the look-ahead bias familiar to quantitative investors. Thus, a workflow can contain properly backtested quantitative components while its agentic behavior remains exposed to a new, separate, LLM-specific form of look-ahead bias.

There are several approaches that can mitigate leakage. Retrieval-augmented generation can restrict the model's supplied evidence to point-in-time information (Zhao and Welsch 2026). Data homogenization can place historical inputs in a common format, while data obfuscation can remove names, dates, ticker symbols ("entity neutering"), and other identifiers that allow the model to recognize a security or episode (Engelberg et al. 2025). Prompts can also instruct the model to reason only from the supplied evidence (Zhou et al. 2023). These methods reduce obvious memory cues but do not remove information encoded in the model's parameters (Sarkar and Vafa 2025). We use a combination of retrieval, homogenization, and obfuscation, but we do not claim that these procedures eliminate look-ahead bias. There is a further issue with LLMs in that even if there is no information leakage, LLM outputs are not perfectly reproducible: even at temperature zero, floating-point arithmetic, batching, and inference hardware can cause nominally identical LLM calls to produce different outputs (Atil et al. 2025; He et al. 2025).

---

[9] Glasserman and Lin (2024) and Levy (2026), among others, document look-ahead bias in financial applications of LLMs. Approaches to mitigate information leakage other than those mentioned in the text include identifying and excluding observations which the model displays evidence of memorization and modifying inference to suppress knowledge associated with post-cutoff information.

One solution is to perform prospective evaluation of agentic systems instead of backtests. Koijen and Levy (2026), for example, optimize model prompts and context on the first half of a quarter and evaluate the resulting system on earnings announcements occurring in the second half. This design permits prompt optimization without requiring point-in-time models because the predicted outcomes are out of sample. We similarly interpret Section 4 as an illustration of how the agents reason and interact, not as evidence of out-of-sample performance. A final note is that the leakage critique does not apply to agentic models in live production environments.

## 5.2 Correlated LLM Errors

If all 21 PC agents and the CIO agent share the same base language model, their errors may be correlated in ways no peer-review protocol fully detects—LLM monoculture risk. The review process partially mitigates this because the agents pursue different optimization objectives with full information, but shared training data and shared failure modes remain a concern. The most robust mitigations are architectural: drawing agents from multiple foundation models to minimize idiosyncratic model errors, and mixing LLM agents with purely deterministic optimizers whose behavior is independent of the language model. Correlated errors are, in our view, the risk most likely to concern a CIO, because they can convert apparent diversification across 21 methods into a single hidden bet.

The fact that the PC agents call deterministic portfolio construction methods mitigates this risk because underlying methods will not yield the same result unless the methods are equivalent. The correlated-error risk arises in the LLM-as-judge layer rather than in the pipeline as a whole. One robustness test is to re-run the pipeline with a heterogeneous mix of foundation models—say, for example, using one model family per PC category.

## *5.3 Model Risk*

Providers update, fine-tune, and deprecate LLMs, and the same prompt can yield materially different outputs across model versions (Chen, Zaharia, and Zou 2023). For an SAA process this is both a reproducibility problem—a recommendation cannot be exactly reproduced once the model behind it has been retired—and an audit problem, since the chain of reasoning that supported a past decision may not be repeatable. Model risk is distinct from look-ahead bias (Section 5.1); even with a perfect point-in-time and reproducible model, version drift can produce non-replicability. Various mitigations include pinning a specific model version for a review cycle, periodically revalidating agent behavior against a fixed test battery, archiving the exact prompts, outputs, and model identifiers behind each decision, and retaining deterministic optimizers as a stable anchor. In our pipeline, each agent is instructed to give plain English rationales behind each decision.

## *5.4 Human Oversight and Automation Surprise*

A new risk compared to traditional human-centered SAA workflows is automation surprise—a phenomenon described by Parasuraman, Sheridan, and Wickens (2000) in which operators lose situational awareness while monitoring an automated system. An investment committee that rubber-stamps the board memo without scrutinizing the CIO’s reasoning risks the same failure modes that occur in aviation and industrial automation. The board memo is written to be readable and flags key assumptions and any deviations from the IPS’s soft targets, but institutional processes must ensure genuine engagement rather than perfunctory oversight—for example, by requiring the committee to record its response to each flagged deviation. Further, the natural-language audit trail produced by the CIO agent, the CRO agent, and every agent in the pipeline is designed to make such engagement feasible and to support explainability and ex-post review.

## 5.5 Governance

The IPS is the load-bearing institutional artifact of our architecture. It is already a legal and regulatory requirement for institutional investors, so repurposing it to govern an agentic pipeline requires no new governance infrastructure: the same document that constrains human portfolio managers can constrain autonomous agents. As Section 4 emphasizes, we distinguish hard constraints (the asset universe and the tracking-error ceiling, which may not be breached) from soft targets (the return, volatility, and drawdown bounds, whose breach must be flagged and justified rather than silently overridden). This design keeps a human in the loop at the point of every consequential deviation and preserves the accountability that Johnson et al. (2017) identify as essential at every level of control abstraction.

## 5.6 Security

There are further risks in the agent interaction protocols. Although the agentic SAA pipeline runs on an internal system, any architecture where agents can invoke tools and modify files—especially for the self-learning meta agent—introduces structural security risk. Techniques like sandboxing, privilege separation for the agents, and other agents acting as auditors can mitigate these risks.

# 6. Discussion: The Role of the Investment Professional

## 6.1 Moving Up the Abstraction Ladder

What is the role of the human in the agentic SAA process?

In the Johnson et al. (2017) framework of increasing human abstraction, a fully autonomous agentic SAA process would have the human writing the IPS (or at least the human is responsible for the IPS)—specifying the asset universe, risk budget, category bounds, and escalation triggers, and the agents execute every subsequent step. Every agent in the pipeline reads the IPS and is constrained by it; the CRO agent explicitly checks IPS compliance for every candidate portfolio; the CIO agent is instructed to comply with the IPS. The IPS is already a legal and regulatory requirement for institutional investors, and repurposing it as the governing document for an agentic pipeline requires no new governance infrastructure; the same document that constrains human portfolio managers can also constrain autonomous agents.

A key feature is that the same agentic SAA workflow can operate at different autonomy levels (Azimbayev et al. 2025) by adjusting the IPS. Tightening the tracking-error bound or adding hard asset-class floors moves the system toward supervisory control; relaxing bounds or removing floors moves it toward full goal-oriented autonomy. The abstraction level is a policy choice, not an architectural constraint.

## 6.2 Autonomous Self-Learning Agents

Our SAA agents already possess the ability to propose new portfolio construction methods (the PC researcher agent), and assess and vote on each other's work (in the PC Strategy Review, see Section 3.4). The agents interact with each other and revise their proposals from other agents' referee reports and the CRO's report. The agents also interpret information on their own (LLM-as-judge). It is possible to extend this further so that agents themselves learn.

The SAA process is not static—the portfolio must be rebalanced regularly, the assumptions underpinning the current target portfolio evaluated (summarized in the last CIO agent's board memo), and the agents' forecasts evaluated. The last includes both the CMAs of the AC agents and the portfolio methodologies of the PC agents.

As part of the ongoing portfolio review and implementation process, we introduce a meta agent that executes a self-improvement cycle after each rebalancing period. First, it computes feedback by comparing the macro agent's and all AC agents' past estimates, signal directions, and regime classifications against realized returns over a rolling three-year window—measuring regime accuracy, cross-sectional rank correlation of expected returns, signal hit rates, and per-method prediction error by asset class and regime. The meta agent analyzes these feedback records to identify systematic weaknesses and researches potential improvements through backtesting and statistical analysis. Then, it decides which improvements to implement and auto-modifies the relevant files: skill and agent prompt descriptions and Python code. Modifications above a materiality threshold must be approved by a human; changes below the threshold can be done by the system. All changes are logged in a structured record that preserves the evidence base, the reasoning, and the exact modifications. We further limit changes to a declared set of files which excludes the IPS and files that involve compliance instructions with the IPS. Since all changes are recorded, they can also be added to the investment committee memo for discussion and ratification—with supervision of the self-improvement loop done the same way as the rest of the pipeline.

This loop is qualitatively different from conventional parameter re-estimation. The meta agent reads its own past performance, reasons about *why* predictions failed, and modifies both computational code and natural-language instructions. For example, a regime-adjusted ERP method that systematically overestimates in late-cycle environments might have its confidence scoring down-weighted, while the CMA judge skill might be updated to apply a larger valuation tilt when CAPE exceeds a threshold—changes that involve both quantitative recalibration and qualitative judgment. The human's role shifts from manually tuning parameters to reviewing the meta agent's change log and setting the constraints that govern the self-improvement process—in the Johnson et al. (2017) framework, supervisory control of the learning process itself.

Of course, because of the long horizons required for SAA evaluations, it will take some time before we can evaluate whether the meta agent's modifications genuinely improve out-of-sample performance.

## *6.3 More Agents*

Our agentic SAA pipeline assigns one agent to a functional role—a single macro agent, a single CRO agent, one agent per asset class—but it is possible to also introduce multiple agents. This is already done by the CIO agent that aggregates and makes decisions over the output of multiple PC agents. Any node in the pipeline can be decomposed into a team of sub-agents coordinated by a supervisor. For example, the macro agent could be replaced by a Chief Economist agent that delegates to specialists—a Fed watcher agent, a fiscal policy agent, a global macro agent, domestic politics agent, geopolitical risk agent, dedicated FX and commodities agents, etc.—and then the Chief Economist agent can aggregate its views through similar deliberation protocols used in the PC strategy review. Similarly, the CRO agent currently produces a single risk report per candidate portfolio; it could instead orchestrate sub-agents for liquidity risk, implementation risk, short-horizon versus long-horizon volatility (each using different covariance estimators produced by yet further agents), coordinate expected shortfall and CVaR decomposition agents, and multiple factor-based stress testing agents, etc.

The guiding principle for when to decompose a single agent into a team is the same one that governs organizational design in human institutions: decompose when the task requires *genuinely distinct expertise* that benefits from independent reasoning before aggregation. A single macro agent can synthesize a regime call when the relevant signals are few and their interactions are well understood; but when the fiscal outlook, monetary stance, and geopolitical environment point in conflicting directions—as they do in a late-cycle regime with simultaneous monetary easing and fiscal expansion—independent specialist agents are more likely to surface the tension than a single agent asked to weigh everything at once (Du et al. 2023). Multiple agents are especially valuable where informed disagreement is productive, not where the computation burden is merely large.

## 7. Conclusion

We describe an agentic architecture for strategic asset allocation (SAA) that coordinates 44 specialized agents through macro analysis, capital market assumptions, portfolio construction, interactions of agents, and CIO analysis. The architecture embeds multi-agent deliberation—peer review, voting, and an adversarial diversifier—into the portfolio construction process, and introduces a meta agent that closes the feedback loop between prediction and realized outcomes so that agents can learn. The Investment Policy Statement (IPS) serves as the operational boundary, analogous to the operational design domain in autonomous driving, ensuring that agents operate within institutionally defined constraints.

There are many open questions. Whether agentic SAA results in better performance than traditional human-centered SAA can only be answered by live performance. How will regulatory frameworks adapt to investment processes in which the analytical workload is executed by autonomous agents governed by an IPS rather than by human analysts governed by organizational and regulatory hierarchy? What is clear is that agents can shift the bottleneck in institutional investing from human bandwidth to human judgment—a shift that elevates rather than diminishes the role of the investment professional.

# Appendix

This appendix collects some more information on implementation referenced in Section 3 for readers interested in how the agents are built. It is not required to follow the investment argument of the main text.

## *A.1. Anatomy of an Agent*

In our agentic architecture, each step of the traditional SAA process is assigned to one or more specialized agents. An agent in this system is defined by four components: a description, a set of scripts, a collection of skills, and a structured output contract.

The ***description*** is a markdown document that defines the agent's role, its position in the pipeline, and its step-by-step workflow. It specifies what inputs the agent consumes, what analyses it performs, and what outputs it produces. This description is read by the LLM at runtime—it is effectively the agent's job description, written in natural language rather than code.

The ***scripts*** are Python programs that handle computation. An agent does not perform arithmetic or optimization in the language model itself; it calls scripts that fetch data from APIs, compute historical statistics, build expected-return models, or run portfolio optimizers. The separation is deliberate: the LLM handles judgment, interpretation, and narrative; the scripts handle computation. This division ensures numerical precision while preserving the agent's ability to reason about results.

The ***skills*** are reusable knowledge modules—each is a folder containing methodology documentation and associated scripts—that multiple agents can share. For example, a `macro-regime` skill contains the regime classification framework (four regimes: expansion, late-cycle, recession, recovery, defined by growth, inflation, monetary policy, and financial conditions scores), the scoring methodology, and the data-fetching script. An `historical-analysis` skill provides functions for loading returns, computing statistics, and generating signals. Skills are the mechanism through which institutional knowledge—valuation frameworks, risk metrics, regime definitions—is encoded and made available to agents.

The ***output contract*** specifies the structured files the agent must produce: JSON files conforming to defined schemas (for machine consumption by downstream agents) and markdown reports (for human review). Every agent produces both quantitative outputs and natural-language analysis, ensuring that the pipeline generates audit trails at every step.

## *A.2. Agent Skills*

We give two examples of agent skill files.

Exhibit A.1 shows a condensed version of the description of the US Large Cap agent which sets CMAs for the S&P 500, illustrating five components: the role description, the required skills, the workflow with script calls including a CMA judge step (see below), asset class-specific guidance, and the structured output contract. Each agent has its own description file with appropriate asset class-specific instructions.

Exhibit A.2 summarizes the CMA judge skill which is used by every equity asset-class agent at the judgment step. It defines seven candidate estimates, a decision framework conditioned on macro regime and valuation context, and the hard constraint that the final estimate must lie within the range of the candidate methods.

**Asset Class Analysis Agent — US Large Cap (S&P 500)**

**Role:** Produces 13-section investment case and capital market assumptions for US Large Cap equities (S&P 500)

**Slug:** us-large-cap | BBG: SPTR Index | Category: US Equity | History: Jan 1990–present

**Macro sensitivity:** Growth (+), Rates (−), Inflation (−), Dollar (+)

**Required skills:** asset-class-report, historical-analysis, signal-generation, equity-analysis, cma-judge, apex-data-financial (fmp, finviz)

**Workflow:**

1. Load macro view (macro-view.json → regime, growth/inflation/policy scores)
2. Historical analysis (returns, vol, drawdowns, correlations vs 12 other asset classes)
3. Valuation & fundamentals (CAPE, P/E, earnings yield, ERP, market breadth)
4. Technical signals (momentum, trend, mean reversion, relative momentum)
5. Sentiment and external data via web search (fund flows, positioning)
6. Build CMA: run 6 methods + auto-blend via cma_methods.py
7. CMA Judge: LLM evaluates 7 candidates, selects final CMA with rationale
8. Scenario analysis (bull/bear from method range) + 13-section report

**Key considerations** (asset-class-specific guidance read at runtime):

- S&P 500 is the benchmark—most investors compare everything against it
- Concentration risk: top 10 stocks may represent 30%+ of index weight
- AI/tech exposure: sector composition heavily tilted toward technology and growth
- Buyback yield: significant CMA component (~1–3% annually)
- Dollar strength impacts multinational earnings (negative for strong dollar on earnings)

---

**Output:** cma_methods.json, cma.json, signals.json, historical_stats.json, scenarios.json, correlation_row.json, analysis.md

**Exhibit A.1**: US Large Cap Agent Configuration (condensed from agent description file)

```
CMA Judge Skill (cma-judge/SKILL.md)
Evaluates multiple CMA methods for an equity asset class and
selects the final expected return estimate.
Inputs:
   cma_methods.json — all method estimates with confidence scores
   signals.json — asset-level macro, technical, valuation signals
   macro-view.json — current regime, growth, inflation, policy
   historical_stats.json — trailing returns, volatility, drawdowns
7 Candidate Methods:
   1. Historical ERP + Rf — long-run realized equity premium
   2. Regime-Adjusted — conditional premium for current macro regime
   3. BL Equilibrium — market-implied return from cap weights
   4. Inverse Gordon — yield + growth from current valuation
   5. Implied ERP — earnings yield as return proxy (CAPE-based)
   6. Survey/Analyst — expert consensus or macro agent view
   7. Auto-Blend — confidence-weighted average of methods 1-6
Judgment Rules:
   Step 1: Assess method dispersion (tight <3pp / moderate 3-6pp / wide >6pp)
   Step 2: Apply regime logic (late-cycle → tilt valuation + regime-adj;
      expansion → auto-blend; recession → regime-adj + BL)
   Step 3: Check valuation context (PE >30x → tilt valuation methods;
      PE <12x → tilt historical + BL; normal → flag if they disagree)
   Step 4: Check signal alignment (confirm or hedge against method spread)
   Step 5: Select — pick one method, define custom weights, or accept blend
Constraint: final estimate MUST be within [min_method, max_method].
```

**Exhibit A.2**: CMA Judge Skill (condensed from cma-judge skill file)